\documentclass[letterpaper]{article} 
\usepackage{aaai2026_camera}  
\usepackage{times}  
\usepackage{helvet}  
\usepackage{courier}  
\usepackage[hyphens]{url}  
\usepackage{graphicx} 
\usepackage{natbib}  
\usepackage{caption} 
\usepackage{algorithm}
\usepackage{algorithmic}

\usepackage{booktabs}
\usepackage[table]{xcolor}
\usepackage{pifont}
\usepackage{amsfonts}
\usepackage{amsmath} 

\usepackage{newfloat}
\usepackage{listings}
\DeclareCaptionStyle{ruled}{labelfont=normalfont,labelsep=colon,strut=off} 
\floatstyle{ruled}
\newfloat{listing}{tb}{lst}{}
\floatname{listing}{Listing}
\title{SemanticVLA: Semantic-Aligned Sparsification and Enhancement \\for Efficient Robotic Manipulation}
\author{
    Wei Li\textsuperscript{\rm 1}, Renshan Zhang\textsuperscript{\rm 1}, Rui Shao\textsuperscript{\rm 1}\thanks{Corresponding authors}, Zhijian Fang\textsuperscript{\rm 1},  Kaiwen Zhou\textsuperscript{\rm 2}, Zhuotao Tian\textsuperscript{\rm 1}, Liqiang Nie\textsuperscript{\rm 1}  \\
}
\affiliations{
    \textsuperscript{\rm 1}Harbin Institute of Technology, Shenzhen\quad 
    \textsuperscript{\rm 2}Huawei Noah’s Ark Lab \\
    liwei2024@stu.hit.edu.cn\quad zhangrenshan@stu.hit.edu.cn\quad shaorui@hit.edu.cn\\
    \textcolor[HTML]{0D47A1}{\url{https://github.com/JiuTian-VL/SemanticVLA}}
}

\begin{document}

\maketitle

\begin{abstract}
Vision-Language-Action (VLA) models have advanced in robotic manipulation, yet practical deployment remains hindered by two key limitations: \textbf{1) perceptual redundancy}, where irrelevant visual inputs are processed inefficiently, and \textbf{2) superficial instruction-vision alignment}, which hampers semantic grounding of actions. 
In this paper, we propose \textbf{SemanticVLA}, a novel VLA framework that performs Semantic-Aligned Sparsification and Enhancement for Efficient Robotic Manipulation. Specifically,
\textbf{1)} To sparsify redundant perception while preserving semantic alignment, \textbf{Semantic-guided Dual Visual Pruner (SD-Pruner)} performs: Instruction-driven Pruner (ID-Pruner) extracts global action cues and local semantic anchors in SigLIP; Spatial-aggregation Pruner (SA-Pruner) compacts geometry-rich features into task-adaptive tokens in DINOv2.
\textbf{2)} To exploit sparsified features and integrate semantics with spatial geometry, \textbf{Semantic-complementary Hierarchical Fuser (SH-Fuser)} fuses dense patches and sparse tokens across SigLIP and DINOv2 for coherent representation.
\textbf{3)} To enhance the transformation from perception to action, \textbf{Semantic-conditioned Action Coupler (SA-Coupler)} replaces the conventional observation-to-DoF approach, yielding more efficient and interpretable behavior modeling for manipulation tasks. Extensive experiments on simulation and real-world tasks show that SemanticVLA sets a new SOTA in both performance and efficiency. SemanticVLA surpasses OpenVLA on LIBERO benchmark by \textbf{21.1\%} in success rate, while reducing training cost and inference latency by \textbf{3.0$\times$} and \textbf{2.7$\times$}. 
\end{abstract}


\begin{figure}[t]
\centering
\includegraphics[width=1\columnwidth]{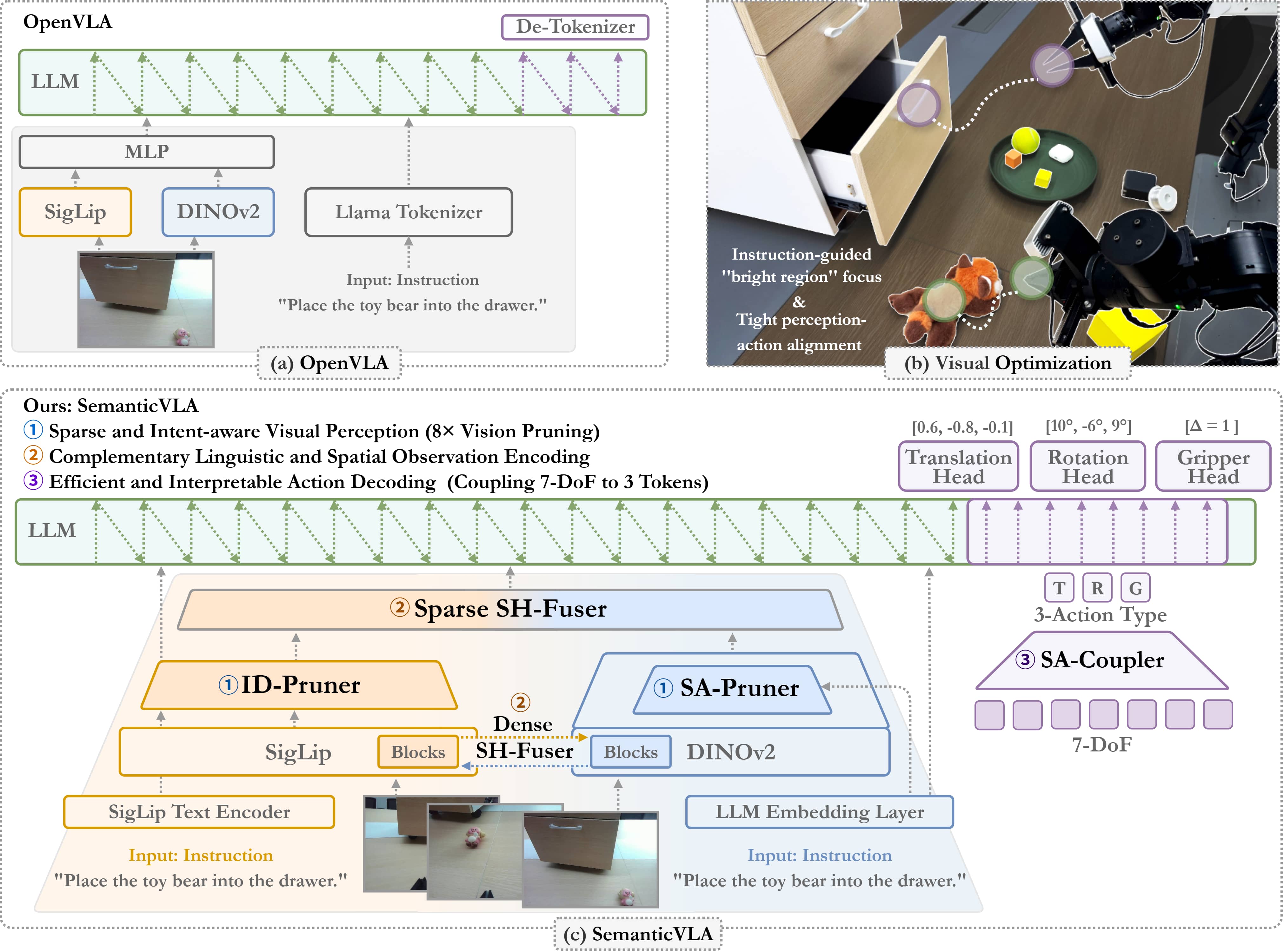}
\caption{Comparison between (a) OpenVLA and (c) SemanticVLA. OpenVLA directly encodes full visual inputs, resulting in redundancy and weak semantic alignment. In contrast, SemanticVLA dynamically performs instruction-guided visual sparsification, establishes tight perception-action correspondence, and accelerates parallel decoding via action type coupling, as illustrated in (b) and (c).}
\label{fig1}
\end{figure}

\section{1~~Introduction}
In recent years, advances in deep learning~\cite{shao2024detecting,shao2023detecting,shao2019multi,wen2023syreanet,zhao2024balf} have driven rapid development of intelligent agents~\cite{li2024optimus, li2025optimus, chen2025SimpAgent, lyu2025puma}. In particular, (VLA)~\cite{Rt-2, openvla, pi0, pi05, Rt-h, li2025cogvla, shao2025large} models have advanced robotic manipulation by enabling end-to-end mapping from language to action using pre-trained Vision-Language Models (VLMs)~\cite{llava, yang2025qwen3, guo2025deepseek, chen2024lion, li2025lion, zhu2025emosym, qiao2025bidirectional}. However, despite their progress, existing VLA models still face critical challenges in deployment, particularly in dynamic and cluttered environments. 
A core bottleneck lies in the lack of semantic-aligned perception and structured action representation, resulting in redundant computation and weak task grounding in complex environments.

This bottleneck is mainly caused by two fundamental limitations:
1) \textbf{Redundancy in visual perception}.  
Prevailing VLA frameworks~\cite{cotvla, hybridvla, univla} adopt generic instruction-agnostic visual encoders~\cite{vit, clip, siglip, dino, dinov2}. They focus on processing all observed pixels uniformly without paying attention to the semantic relevance of the instructions. As a result, background clutter, task-irrelevant distractors, and environmental noise are encoded indiscriminately, leading to excessive computational cost and diluted attention to task-critical cues.
2) \textbf{Superficiality in instruction-vision semantic alignment.} 
Most VLA models~\cite{molevla, deervla,openvla, pi0} rely solely on generic cross-modal alignment with large language models. This superficial alignment struggles to capture complex semantic relations in robotics and thus failing to capture fine-grained visual compositionality. Consequently, this significantly limits the VLA's ability to identify global action cues, local semantic anchors, and the structured instruction-spatial dependencies in robotic manipulation tasks.

To address these challenges, we propose \textbf{SemanticVLA}, a novel framework designed to perform Semantic-Aligned Sparsification and Enhancement for efficient and interpretable robotic manipulation (Fig.~\ref{fig1}). SemanticVLA hinges on three-level complementary semantics: \textbf{1)} \textbf{instruction-level linguistic intent semantics} conveyed by task prompts; \textbf{2)} \textbf{vision-level spatial semantics} describing objects and their layout; and \textbf{3)} \textbf{control-level action semantics} governing translation, rotation, and gripper state. 

Specifically, SemanticVLA unifies sparsification and enhancement aligned to these semantics through three integrated modules: 
\textbf{1) Semantic-guided Dual Visual Pruner (SD-Pruner).} Exploiting encoder specialization—SigLIP~\cite{siglip} for instruction grounding and DINOv2~\cite{dinov2} for spatial geometry, SD-Pruner independently prunes each encoder to retain the most task-relevant evidence under occlusion and noise. 
\textbf{i) Instruction-driven Pruner (ID-Pruner) for SigLIP.} We compute instruction–image cross-modality similarity to derive token importance scores, enabling two complementary paths: 
Vision-to-Language Mapping preserves global action cues from complete instruction inputs, resolving the ``know the goal but not the steps'' issue. Language-to-Vision Filtering enhances local semantic anchors from complete visual inputs, mitigating the ``can’t do what you can’t see'' problem. Together, they form a robust and sparsified perception pipeline that retains essential visual-language-action alignment under occlusion and noise. 
\textbf{ii) Spatial-aggregation Pruner (SA-Pruner) for DINOv2.} Inspired by the register design in previous work~\cite{zhang2025falcon}, SA-Pruner aggregates DINOv2 features into compact, geometry-rich tokens, further modulated via FiLM~\cite{perez2018film} to reflect instruction relevance, thereby complementing SigLIP’s semantics.
\textbf{2) Semantic-complementary Hierarchical Fuser (SH-Fuser)}. To enhance coherence between spatial vision and task intent, SH-Fuser performs a two-stream fusion that spans the entire visual encoding stage. Dense-Fuser exchanges patch-level information between corresponding blocks of SigLIP and DINOv2, while Sparse-Fuser merges salient tokens produced by ID-Pruner and SA-Pruner. This hierarchical design propagates complementary cues early and late in the stack, producing a unified representation that is both semantically grounded and geometrically faithful.
\noindent\textbf{3) Semantic-conditioned Action Coupler (SA-Coupler).} To enable efficient and interpretable observation-to-action mappings, SA-Coupler adopts a structured formulation that explicitly maps perception representations to semantic action types. Joint control is modulated based on them accordingly, which further enhances both the efficiency and interpretability of action decoding.

We evaluate SemanticVLA on extensive simulation and real-world manipulation tasks, showing improved task success with reduced perceptual redundancy, enhanced instruction reasoning, and lower computational cost over SOTA baselines. Our contributions are summarized as follows:

\begin{itemize}
    \item We propose \textbf{SD-Pruner}, which jointly prunes SigLIP and DINOv2 encoders via instruction-aware token filtering and geometry-aware aggregation via \textbf{ID-Pruner} \& \textbf{SA-Pruner}, significantly pruning redundant perception.
    \item We propose \textbf{SH-Fuser}, a two-stream fusion module that integrates dense patch features and sparse semantic tokens across SigLip and DINOv2, enhancing instruction semantics and spatial structure alignment.
    \item We design \textbf{SA-Coupler} to enable a more intuitive and efficient mapping from sparsified perception to semantic action types.
    \item Extensive experiments on standard VLA benchmarks and real-world robot deployments demonstrate that SemanticVLA achieves SOTA performance and efficiency.
\end{itemize}

\section{2~~Related Work}

\paragraph{Robot Manipulation with Lightweight Models.} These models \cite{diff, lee2024behavior, star, lv2025spatial} typically excels in deterministic and real-time control. However, these approaches heavily rely on pre-defined objects and environments, lacking the semantic generalization capabilities necessary for open-world settings.

\paragraph{Two Branches of VLM-based VLA Models.} VLA can be broadly categorized into two types. Monolithic architectures \cite{Rt-2, openvla, zhao2025vlas, spatialvla}, which maintain causal and semantically consistent multi-step reasoning, making them well-suited for open-ended environments. Nevertheless, their reliance on autoregressive action decoding introduces substantial efficiency bottlenecks. Hierarchical expert models \cite{pi0, cogact, wen2024diffusionvla, pi05, bjorck2025gr00t, shukor2025smolvla} which leverage diffusion or flow-matching mechanisms for high-frequency action prediction. While effective, these models suffer from a disconnect between the VLM and action experts, thereby underutilizing the reasoning capacity of the VLM.

\paragraph{Efficient VLA Modeling Approaches.} Efficient VLA models generally fall into three categories: 1) Algorithmic strategies like FAST \cite{fast}, which merges action bins using discrete cosine transform (DCT), and PD-VLA~\cite{pdvla} or OpenVLA-OFT~\cite{oft}, which employ parallel decoding, aim to accelerate inference. 2) Architectural innovations include RoboMamba's Mamba backbone \cite{liu2024robomamba}, Deer-VLA's multi-exit design \cite{deervla}, and MoLe-VLA's dynamic layer-skipping mechanism \cite{molevla}, all of which reduce redundant computation. 3) Compression-based approaches such as NORA \cite{hung2025nora} and BitVLA \cite{wang2025bitvla} focus on downsizing models while retaining task performance. While these strategies enhance internal efficiency, they often neglect the alignment between visual inputs and instruction semantics—an essential component for embodied creativity, which is inherently visual.

\begin{figure*}[t]
\centering
\includegraphics[width=1\textwidth]{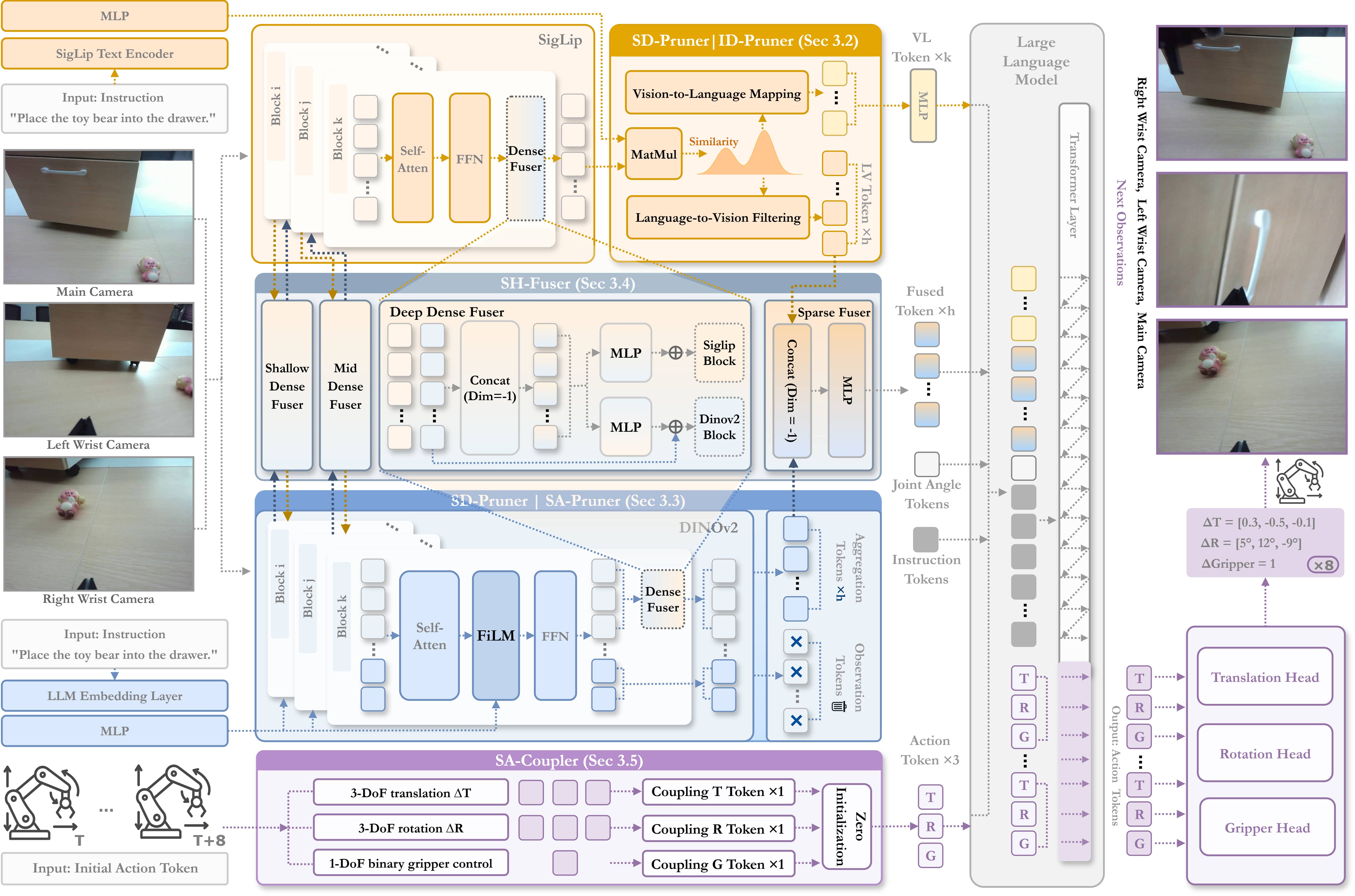}
\caption{\textbf{Overview of the SemanticVLA Framework.} Observations are processed through two parallel pathways: instruction-aware encoding via SigLIP-based Instruction-driven Pruner and spatial-aware encoding via DINOv2-based Spatial-aggregation Pruner, tightly fused through a shared Semantic-complementary Hierarchical Fuser. Action inputs are initialized via Semantic-conditioned Action Coupler to optimize the sparsified perception to action type transition in large language model.}
\label{fig2}
\end{figure*}

\section{3~~SemanticVLA}

\subsection{3.1~~Proposed Framework}
We define the input context as $\mathbf{X} = \{\mathcal{V}, \mathbf{q}, \mathbf{\ell}\}$, where $\mathcal{V}$ denotes the real-time visual observation, $\mathbf{q}$ represents the robot's current proprioceptive state (e.g., joint angles and end-effector pose), and $\mathbf{\ell}$ is the provided natural language instruction. The model predicts a chunk of $K$ future actions $\mathbf{A} = [\mathbf{a}_0, \mathbf{a}_1, \dots, \mathbf{a}_{K-1}] \in \mathbb{R}^{(K \times D) \times d}$, where $D$ signifies the dimensionality of each atomic action vector (e.g., $D=7$ for 3-DoF translation, 3-DoF rotation, and gripper control). 

As shown in Fig.~\ref{fig2}, SemanticVLA processes $\mathcal{V}$ through two parallel pathways: 1) a SigLIP-based visual encoder whose outputs are sparsified via \textbf{ID-Pruner} according to instruction guidance; and 2) a DINOv2-based spatial encoder that captures dense geometric features via \textbf{SA-Pruner}. These two streams are hierarchically integrated through the \textbf{SH-Fuser} to produce a task-relevant representation $\mathbf{Z}$. And then, $\mathbf{Z}$ is concatenated with $\mathbf{\ell}$, $\mathbf{q}$, and $K$ learnable action placeholders, constructing the input for parallel decoding:
\begin{equation}
    \mathbf{\tilde{X}} = [\mathbf{Z}, \mathbf{q}, \mathbf{\ell}, \mathbf{0}_0, \mathbf{0}_1, \dots, \mathbf{0}_{K-1}] 
\end{equation}
where, following \textbf{SA-Coupler}, each placeholder $\mathbf{0}_i=\{\mathbf{t}_i^0,\mathbf{r}_i^0,\mathbf{g}_i^0\}\in\mathbb{R}^{3\times d}$ explicitly separates translation, rotation, and gripper tokens while jointly encoding the full 7-DoF atomic action vector. Lastly, a bidirectional decoding process $f_{\parallel}(\cdot)$ performs a single forward pass on $\mathbf{\tilde{X}}$ to concurrently generate all $K$ future actions: $\mathbf{A} = f_{\parallel}(\mathbf{\tilde{X}})$. 
This structured pipeline enables SemanticVLA to achieve efficient and instruction-aware manipulation by tightly integrating semantic sparsification, hierarchical fusion, and compositional action modeling within a unified architecture.

\begin{figure*}[t]
\centering
\includegraphics[width=1.0\textwidth]{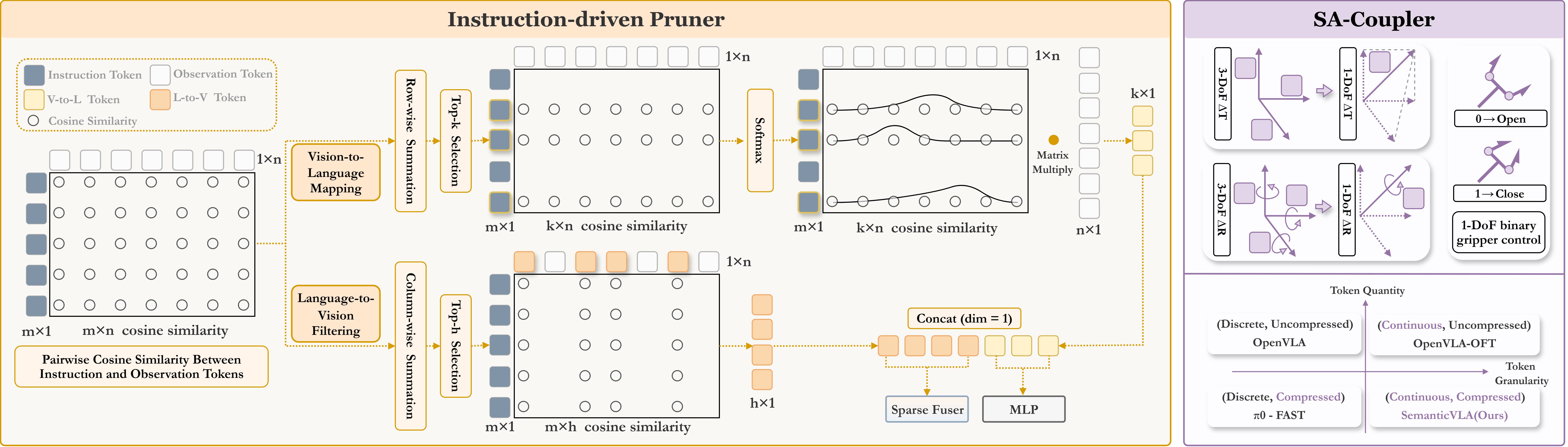}
\caption{Illustration of Instruction-driven Pruner for SigLIP (left) and Semantic-conditioned Action Coupler (right).}
\label{fig3}
\end{figure*}

\subsection{3.2~~ID-Pruner for SigLIP}
ID-Pruner executes a dual pruning mechanism between visual tokens \( \{\mathbf{v}^{Sig}_i\}_{i=1}^N \in \mathbb{R}^{N \times d_v^{Sig}} \), extracted via SigLIP encoder, and instruction embeddings \( \{\mathbf{l}^{Sig}_{j}\}_{j=1}^M \in \mathbb{R}^{M \times d_l^{Sig}} \) generated by the SigLip text encoder. This process operates through two pathways: Vision-to-Language Mapping for global action cues, and Language-to-Vision Filtering for local semantic anchors, as shown in Fig.~\ref{fig3} (left).

\paragraph{Step 1:} Cosine Similarity Matrix Construction. Each instruction token \( \mathbf{l}^{Sig}_{j} \) is projected into the visual token space using a transformation matrix \( \mathbf{W}_l \in \mathbb{R}^{d_v^{Sig} \times d_l^{Sig}} \), followed by cosine similarity computation with each visual token:
\begin{equation}
\textstyle \mathbf{S}_{ij} = \text{sim}\left( \mathbf{W}_l \cdot \mathbf{l}^{Sig}_{j},\ \mathbf{v}^{Sig}_i \right)= \frac{(\mathbf{W}_l \cdot\mathbf{l}^{Sig}_j)^\top \mathbf{v}^{Sig}_i}{\|\mathbf{W}_l \cdot\mathbf{l}^{Sig}_j\| \cdot \|\mathbf{v}^{Sig}_i\|}
\end{equation}
where \( \mathbf{S} \in \mathbb{R}^{N \times M} \) denotes the visual-instruction similarity matrix, and captures the fine-grained relevance between every visual patch and every word in the instruction.

\paragraph{Step 2:} Vision-to-Language Mapping. 
This pathway identifies key instruction tokens (e.g., target nouns, action verbs) and pinpoint their visual correspondents in the scene. We compute a saliency score $s_j^{\text{VL}}$ for each instruction token $\mathbf{l}^{Sig}_j$ by aggregating its similarities across all visual tokens.
\begin{equation}
\textstyle s_j^{\text{VL}} = \sum_{i=1}^N \mathbf{S}_{ij}, \quad \mathcal{I}_{\text{top}-k} = \arg\max_k \left\{ s_j^{\text{VL}} \right\}
\end{equation}
where \( \mathcal{I}_{\text{top}-k} = \{i_1,\dots,i_k\} \) represents the indices of the top-$k$ most salient instruction tokens, typically capturing the task's key elements. Subsequently, for each selected instruction token $\mathbf{l}^{Sig}_p$ (where $p \in \mathcal{I}_{\text{top}-k}$), it is aggregated using softmax-normalized weights to form a global cue vector:
\begin{equation}
\textstyle \alpha_{j_p} = \frac{\exp\left( s_{j_p}^{\text{VL}} \right)}{\sum_{q=1}^k \exp\left( s_{j_p}^{\text{VL}} \right)}, \quad 
\mathbf{v}^{Sig}_{\text{p}} = \sum_{p=1}^k \alpha_{j_p} \cdot \mathbf{v}^{Sig}_{j_p}
\end{equation}
Finally, we obtain\( \mathcal{V}^{\text{VL}} = \{\mathbf{v}^{Sig}_p | p \in \mathcal{I}_{\text{top}-k}\}  \in \mathbb{R}^{k \times d_v} \) as the instruction-aware global action cue feature.

\paragraph{Step 3:} Language-to-Vision Filtering. 
In contrast to the VL Mapping path, this pathway aims to identify and preserve visual regions that are most relevant to the local semantic anchors of overall instruction. We compute a comprehensive relevance score $s_i^{\text{LV}}$ for each visual token $\mathbf{v}^{Sig}_i$ by aggregating its similarities across all instruction tokens. This evaluates the total ``response strength" of each visual region to the instruction as a whole: 
\begin{equation}
\textstyle s_i^{\text{LV}} = \sum_{j=1}^M \mathbf{S}_{ij}, \quad \mathcal{I}_{\text{top}-h} = \arg\max_h \left\{ s_i^{\text{LV}} \right\}
\end{equation}
where \( \mathcal{I}_{\text{top}-h} = \{i_1,\dots,i_h\} \) represents the indices of the top-$h$ visual tokens. We select the Top-h visual tokens with the highest scores, forming a sparse yet critical visual subset \( \mathcal{V}^{\text{LV}} = \{\mathbf{v}_q | q \in \mathcal{I}_{\text{top}-h}\} \). This step effectively filters out background noise and irrelevant distractors, achieving an initial focus on key regions with local semantic anchors. 

\paragraph{Step 4.} The final output of ID-Pruner is the union of the two pruned sets of visual tokens: \(  \mathcal{V}^{\text{VL}} \cup \mathcal{V}^{\text{LV}} \in \mathbb{R}^{(k+h) \times d_v^{Sig}} \). This dual-path design balances global action cues (to avoid misinterpreting manipulation details) with local semantic anchors (to avoid missing key regions). It achieves efficient visual compression while maximally preserving the essential vision-language-action joint information.

\begin{table*}[t]
\caption{\textbf{Simulation Results.} Comparison of task success rates (SR) and ranks (RK) across four suites in the LIBERO benchmark.}
    \label{tab:libero-p}
    \centering
    \footnotesize
    \setlength{\tabcolsep}{6.5pt}  
    \begin{tabular}{l|cc|cc|cc|cc|cc}
    \toprule
    \textbf{Method} & \multicolumn{2}{c|}{\textbf{Spatial}} & \multicolumn{2}{c|}{\textbf{Object}} & \multicolumn{2}{c|}{\textbf{Goal}} & \multicolumn{2}{c|}{\textbf{Long}} & \multicolumn{2}{c}{\textbf{Overall}} \\ & SR $\uparrow$ & RK $\downarrow$ & SR $\uparrow$ & RK $\downarrow$ & SR $\uparrow$ & RK $\downarrow$ & SR $\uparrow$ & RK $\downarrow$ & SR $\uparrow$ & RK $\downarrow$ \\
    \midrule
    Octo fine-tuned~\textit{[RSS'24]}~\cite{octo}          & 78.9 & 10 & 85.7 & 10 & 84.6 &  8 & 51.1 & 10 & 75.1 & 10 \\
    $\pi_0$ fine-tuned~\textit{[arXiv'24]}~\cite{pi0}      & 96.8 &  4 & \underline{98.8} &  \underline{2} & \underline{95.8} &  \underline{3} & 85.2 &  6 & 94.2 &  6 \\
    OpenVLA~\textit{[CoRL'24]}~\cite{openvla}              & 84.7 &  9 & 88.4 & 9 & 79.2 &  9 & 53.7 &  9 & 76.5 &  9 \\
    OpenVLA-OFT~\textit{[arXiv'25]}~\cite{oft}             & \underline{97.6} &  \underline{2} & \underline{98.4} &  \underline{3} & \textbf{97.9} &  \textbf{1} & \underline{94.5} &  \underline{2} & \underline{97.1} &  \underline{2} \\
    STAR~\textit{[ICML'25]}~\cite{star}                                & 95.5 &  5 & 98.3 &  5 & 95.0 &  5 & 88.5 &  5 & 94.3 &  5 \\
    CoT-VLA~\textit{[CVPR'25]}~\cite{cotvla}                  & 87.5 &  8 & 91.6 &  7 & 87.6 &  7 & 69.0 &  7 & 83.9 &  7 \\
    PD-VLA\dag~\textit{[arXiv'25]}~\cite{pdvla}                & 95.5 &  5 & 96.7 &  6 & 94.9 &  6 & \underline{91.7} &  \underline{3} & 94.7 &  4 \\
    SpatialVLA~\textit{[RSS'25]}~\cite{spatialvla}         & 88.2 &  7 & 89.9 &  9 & 78.6 & 10 & 55.5 &  8 & 78.1 &  8 \\
    \toprule
    \textbf{SemanticVLA-Lite } & \underline{97.0} & \underline{3} & \underline{98.4} & \underline{3} & 95.4 &  4 & \underline{92.4} & \underline{3} & \underline{95.8} & \underline{3} \\
    \textbf{SemanticVLA } &\textbf{98.6} &\textbf{1} & \textbf{99.6} & \textbf{1} & \underline{97.6} &  \underline{2} &\textbf{94.8} & \textbf{1} & \textbf{97.7} & \textbf{1} \\
    \bottomrule
    \end{tabular}
\end{table*}

\subsection{3.3~~SA-Pruner for DINOv2}

Parallel to the SigLIP-based pruning branch, we employ the DINOv2-based SA-Pruner to extract dense spatial representations from the observation tokens \( \mathcal{V}^{Din}  \in \mathbb{R}^{N \times d_v^{Din}} \). To facilitate spatial aggregation, a set of zero-initialized aggregation tokens 
\(\mathcal{V}^{\text{Agg}} \in \mathbb{R}^{(N/8) \times d_v^{Din}}\) is appended to $\mathcal{V}^{Din}$. As a self-supervised model, DINOv2 excels at capturing fine-grained spatial structure and geometric details. This makes it an ideal source of dense spatial features to complement the sparse, object-centric features provided by ID-Pruner. 

To align dense spatial features with task semantics, we introduce lightweight instruction modulation via a FiLM layer. A pooled instruction representation \( \mathbf{\bar{\ell}}^{Din} \) is passed to produce affine transformation parameters (scale \( \mathbf{\gamma} \) and shift \( \mathbf{\beta} \)):
\begin{equation}
(\mathbf{\gamma}, \mathbf{\beta}) = \text{FiLM}(\mathbf{\bar{\ell}}^{Din}) \in \mathbb{R}^{d_v^{Din} \times 2}
\end{equation}
These parameters are subsequently applied to the concatenated \(\mathcal{V}^{Din} \cup \mathcal{V}^{\text{Agg}}\),yielding the modulated representation:
\begin{equation}
(\mathcal{V}^{Din} \cup \mathcal{V}^{\text{Agg}})' = (\mathbf{1} + \mathbf{\gamma}) \odot \text{Attn}(\mathcal{V}^{Din} \cup \mathcal{V}^{\text{Agg}}) + \mathbf{\beta}
\end{equation}
where \( \odot \) denotes element-wise multiplication. This formulation allows spatial features to be dynamically adjusted based on task context, enabling aggregation onto \( \mathcal{V}^{\text{Agg}} \), while maintaining semantic relevance. The resulting representations of SA-Pruner are structurally aligned with the outputs of ID-Pruner, facilitating effective cross-modal fusion.

\subsection{3.4~~SH-Fuser cross SigLip \& DINOv2}
SH-Fuser hierarchically integrates sparse semantic features from ID-Pruner with dense geometric-rich features from SA-Pruner through dynamic, layer-wise modulation rather than simple late-stage concatenation, as shown in Fig.~\ref{fig2}.

\paragraph{Dense-Fuser.} This module is inserted between multiple Transformer blocks at different depths (for example, once in superficial, intermediate, and deep layers). This hierarchical integration makes sure that semantic cues (from SigLIP) are enhanced with corresponding spatial-geometric priors (from DINOv2) at each stage, thus enabling the synergistic enhancement of the two complementary visual streams. For the $b-th$ block, the fusion operation is defined as:
\begin{equation}
 \mathcal{V}^{Fusion}_{b} = \text{MLP}(\text{Concat}(\mathcal{V}^{Sig}_{b}, \mathcal{V}^{Din}_{b}))\in \mathbb{R}^{N \times d_v}
\end{equation}

\paragraph{Sparse-Fuser.} 
At the final stage, Sparse-Fuser merges the salient outputs from both pruning paths, where $\mathcal{V}^{LV}$ originates from the ID-Pruner and $\mathcal{V}^{Agg}$ from the SA-Pruner, forming a compact representation: 
\begin{equation}
 \mathbf{Z}^{Fusion}= \text{MLP}(\text{Concat}(\mathcal{V}^{LV}, \mathcal{V}^{Agg}))\in \mathbb{R}^{h \times d_l}
\end{equation}
The Semantic-complementary Hierarchical Fuser reduces visual tokens by $8$–$16\times$ while preserving discriminative representations. This design not only greatly improves computational efficiency (through token pruning) but also enhances task performance by making use of the complementary advantages of semantic grounding and geometric precision.

\subsection{3.5~~Semantic-conditioned Action Coupler}
Building on Sections 3.1-3.3, we construct the visual input to the LLM as a composite token set $\mathbf{Z} = \mathbf{Z}^\text{VL} \cup \mathbf{Z}^\text{Fusion}$ $~(\mathbf{Z}^\text{VL} = \text{MLP}(\mathcal{V}^{\text{VL}})\in \mathbb{R}^{h \times d_l})$, which integrates both semantic and spatially-aligned visual information. 

To optimize the mapping from vision to action, we depart from the conventional VLA formulation that discretizes 7-DoF actions into 7 independent binned tokens, each corresponding to a single DoF. Instead, as illustrated in Fig.~\ref{fig2} and Fig.~\ref{fig3} (right), we introduce a novel Semantic-conditioned Action Coupler that restructures the visual-to-action pipeline in a more structured and intuitive manner:

\paragraph{Token-Level Semantic Alignment.}  
Each of the three fundamental motion primitives (3-DoF translation, 3-DoF rotation, and 1-DoF gripper control) is represented by a single token, enabling unified and semantically coherent modeling of atomic action types:
    \begin{equation}
     \mathbf{0}_i = \{\mathbf{t}_i^0, \mathbf{r}_i^0, \mathbf{g}_i^0\} \in \mathbb{R}^{3 \times d_l}
    \end{equation}

\paragraph{Head-Level Modularity for Action Prediction.}  
The input sequence $[\mathbf{Z}, \mathbf{q}, \mathbf{\ell}, \mathbf{0}_0, \mathbf{0}_1, \dots, \mathbf{0}_{K-1}]$ is fed into the $f_{\parallel}(\cdot)$, yielding an updated action representation $\mathbf{h}_i = \{\mathbf{t}_i^h, \mathbf{r}_i^h, \mathbf{g}_i^h\} \in \mathbb{R}^{3 \times d_l}$. During token-to-value decoding, we design three prediction heads, each specialized for one action type, to directly regress continuous motion parameters: 
    \begin{equation}
     \mathbf{d}_{i,u} = \mathbf{W}_u \mathbf{h}_i + \mathbf{b}_u, \quad \mathbf{W}_u \in \mathbb{R}^{D_u \times d_{l}}, \mathbf{b}_u \in \mathbb{R}^{D_u}
    \end{equation}
where $u \in \{\text{trans},\text{rot},\text{grip}\}$, and $D_u$ denotes the dimensionality of degrees of freedom for action type $u$. 

\begin{table*}[t]
\caption{\textbf{Efficiency Results in Simulation.} SemanticVLA-Lite and SemanticVLA achieve the highest efficiency and the best trade-off between efficiency and performance, respectively. ``\dag'' denotes our reproduced results. Z and H indicate the number of visual input tokens and initialized action tokens. Throughput refers to the number of actions predicted per second during inference. Efficiency results on real-world tasks conducted on the AgileX Cobot Magic platform are reported in \textbf{Appendix~D.3}.}
    \label{tab:efficiency}
    \centering
    \footnotesize
    \setlength{\tabcolsep}{5.2pt}
    \begin{tabular}{l|cccccc}
    \toprule
    \textbf{Method} & \textbf{Z \& H tokens} $\downarrow$ & \textbf{FLOPs} $\downarrow$ & \textbf{Training Cost} $\downarrow$ & \textbf{Latency} $\downarrow$ & \textbf{Throughput} $\uparrow$ & \textbf{LIBERO SR} $\uparrow$ \\
    \midrule
    OpenVLA\dag~\cite{openvla} & 256 \& 7 & 8.48 T  & 11.7 h  & 0.240 s   & 4.2 Hz & 76.5\%  \\
    OpenVLA-OFT\dag~\cite{oft} & 256 \& 7 & 8.45 T  & 12.3 h  & 0.134 s  & 59.7 Hz  & 97.1\%  \\
    PD-VLA\dag~\cite{pdvla} & 256 \& 7  &8.48 T  & 11.7 h     & 0.143 s  & 55.9 Hz   & 94.7\%  \\
    \midrule
    \textbf{SemanticVLA-Lite} & 16 \& 3 & 1.93 T   & 3.6 h  & 0.087 s  & 92.0 Hz  & 95.8\% \\
    \textbf{SemanticVLA}  & 32 \& 3 & 2.37 T  & 3.9 h    & 0.089 s  & 89.9 Hz  & 97.7\% \\
    \bottomrule
    \end{tabular}
\end{table*}

\begin{table*}[t]
\caption{\textbf{Real-World Results.} Comparison of AgileX Cobot Magic task and subtask success rates. Decimal values arise from averaging over 15 trials. The overall SR reflects only the final subtask success rates. ``\dag'' denotes our reproduced results.}
\label{tab:aloha}
\centering
\footnotesize
\setlength{\tabcolsep}{5pt}
\begin{tabular}{l|cc|ccc|ccc|c}
\toprule
\textbf{Method} & \multicolumn{2}{c|}{\textbf{Object Placement}} & \multicolumn{3}{c|}{\textbf{Drawer Manipulation}} & \multicolumn{3}{c|}{\textbf{T-shirt Folding}} & \textbf{Overall} \\
& Bear $\rightarrow$Plate & +Raccoon$\rightarrow$Bowl & Open  & +Place & +Close & Step 1 & +Step 2 & +Step 3 & SR\\
\midrule
VQ-BeT~\cite{be}* & 5/10 & 3/10 & 4/10 & 3/10 & 1/10 & - & - & -& 20.0\% \\
QueST~\cite{quest}* & 6/10 & 4/10 & 3/10 & 1/10 & 0/10 & - & - & -& 20.0\% \\
STAR*~\cite{star} & 8/10 & 6/10 & 6/10 & 4/10 & 3/10 & - & - & -& 45.0\% \\
PD-VLA\dag & 7.3/10 & 6.7/10 & 6.0/10 & 5.3/10 & 4.7/10 & 6.7/10 & 5.3/10 & 4.0/10 & 51.1\%\\
OpenVLA-OFT\dag & 8.0/10 & 6.7/10 & 7.3/10 & 6.7/10 & 5.3/10 & 6.7/10 & 6.0/10 & 4.7/10& 55.6\%\\
\midrule
\textbf{SemanticVLA-Lite} & 8.0/10 & 7.3/10 & 8.0/10 & 6.7/10 & 5.3/10 & 8.7/10 & 8.0/10 & 6.7/10 & 62.2\%\\
\textbf{SemanticVLA }& \textbf{9.3/10} & \textbf{9.3/10} & \textbf{8.7/10} & \textbf{7.3/10} & \textbf{6.0/10} & \textbf{9.3/10} & \textbf{8.7/10} & \textbf{8.0/10}& \textbf{77.8\%} \\
\bottomrule
\end{tabular}
\end{table*}

\begin{figure*}[t]
\centering
\includegraphics[width=1.0\textwidth]{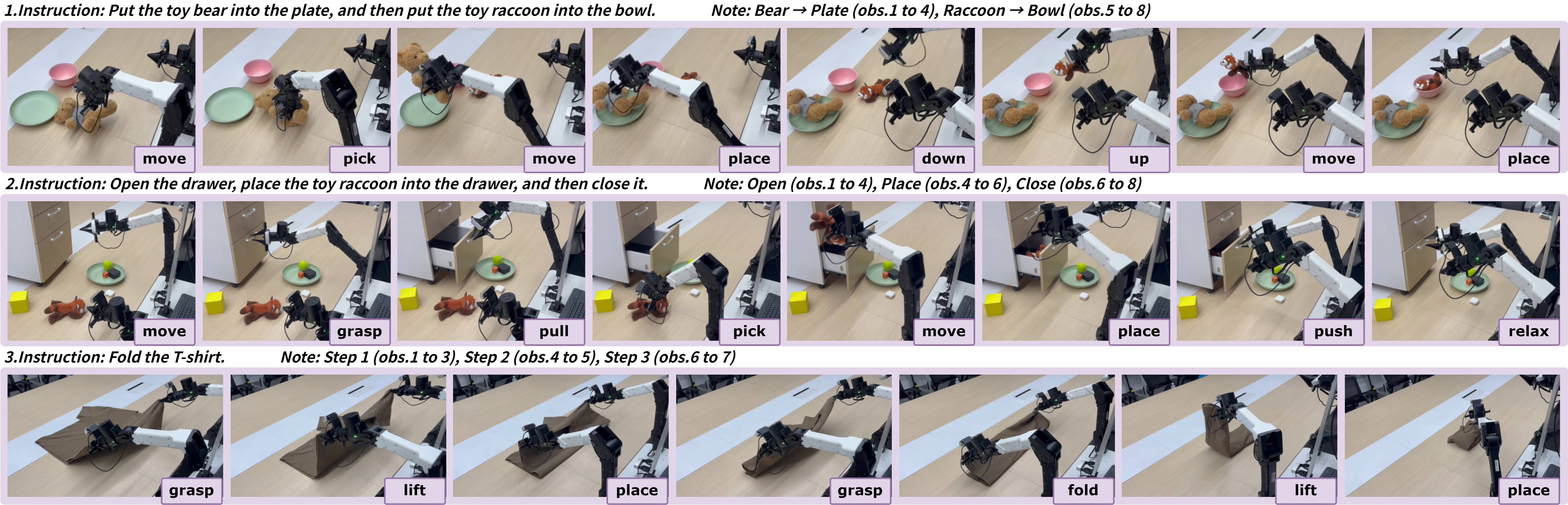}
\caption{Visualization of SemanticVLA's manipulation process on three long-horizon real-world tasks, showing key execution-stage observations.}
\label{vis1}
\end{figure*}

\begin{figure*}[t]
\centering
\includegraphics[width=1.0\textwidth]{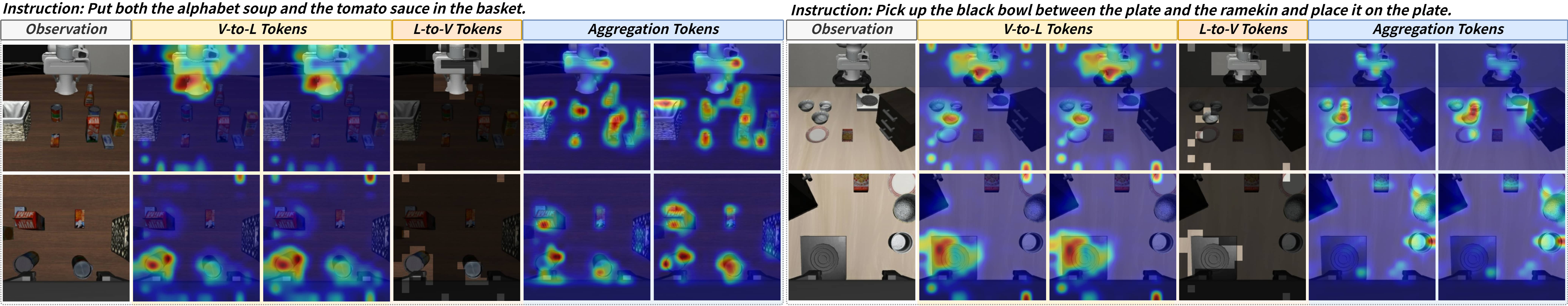}
\caption{Visualization of 1) attention maps from V-to-L tokens to observation patches, capturing global action cues; 2) selected L-to-V token sets, where highlighted patches represent local semantic anchors; and 3) attention maps from aggregation tokens to patches, reflecting spatial features that complement ID-Pruner via HF-Fuser.}
\label{vis2}
\end{figure*}

\section{4~~Experiments}
\subsection{4.1~~Experiment Settings}
\paragraph{Implementation details.} All experiments are conducted on 8× A800 (80GB) GPUs.

\paragraph{Simulation \& Real-World Setup.} Simulation evaluations are performed on the LIBERO benchmark~\cite{libero}, which includes four task suites: Spatial, Object, Goal, and Long, each with 500 human-teleoperated demonstrations. Real-world experiments are conducted on the AgileX Cobot Magic platform~\cite{aloha}, covering object placement, drawer manipulation, multi-step deformable tasks, and two additional scenarios, with 60, 60, 45, and 105 human-teleoperated demonstrations, respectively.

\paragraph{Baselines.} We compare against SOTA baselines on LIBERO, including OpenVLA, Octo and $\pi_0$. Efficiency comparisons are conducted under identical settings with OpenVLA and its accelerated variants, OpenVLA-OFT (the best-performing baseline) and PD-VLA. The top-3 performing models are further validated in real-world deployments.

\subsection{4.2~~Overall Performance \& Efficiency}
\paragraph{Simulation Experimental Analysis.} We evaluate SemanticVLA on four suites, which cover spatial reasoning, object generalization, goal understanding, and long-horizon tasks. 
\begin{enumerate}
    \item[1)] As shown in \textbf{Tab.~\ref{tab:libero-p}}, SemanticVLA achieves the highest success rate (97.7\%, rank 1), consistently outperforming recent SOTA methods through its Semantic-Aligned Sparsification and Enhancement design. SemanticVLA-Lite attains 95.8\% (rank 3) with reduced complexity, highlighting the robustness and scalability. 
    \item[2)] As reported in \textbf{Tab.~\ref{tab:efficiency}}, SemanticVLA achieves superior training (FLOPs, cost) and inference (latency, throughput) efficiency while utilizing only 1/16 or 1/8 of visual inputs and 3/7 of action representations. It significantly outperforms OpenVLA and other efficient baselines. 
\end{enumerate}

\begin{table}[t]
\caption{\textbf{Ablation on SD-Pruner.} SigLip and DINOv2 adopt different pretraining paradigms, with ID-Pruner and SA-Pruner, respectively.}
    \label{tab:SD-Pruner}
    \centering
    \footnotesize
    \setlength{\tabcolsep}{3pt}  
    \begin{tabular}{cc|ccccc}
    \toprule
    \textbf{SigLip} & \textbf{DINOv2} & \textbf{Spatial} & \textbf{Object} & \textbf{Goal} & \textbf{Long} & \textbf{Overall} \\
    \midrule
    ID-Pruner& ID-Pruner& 96.6  & 97.4 & 83.8 & 89.6 & 91.9\\
    SA-Pruner & SA-Pruner & 96.2 & 96.8 & 95.2 & 90.2 & 94.6\\
    SA-Pruner & ID-Pruner& 96.2 & 98.0 & 94.4 & 91.4 & 95.0\\
    \midrule
    \textbf{ID-Pruner}& \textbf{SA-Pruner} & \textbf{98.2} & \textbf{99.0} & \textbf{97.2} & \textbf{93.8} & \textbf{97.1}\\
    \bottomrule
    \end{tabular}
\end{table}

\paragraph{Real-world Experimental Analysis.} We evaluate SemanticVLA on multiple real-world tasks to assess its effectiveness and efficiency.
\begin{enumerate}
    \item[1)] As shown in \textbf{Tab.~\ref{tab:aloha}}, SemanticVLA achieves a success rate of up to 77.8\% on three challenging long-horizon tasks (Object Placement, Drawer Manipulation, and T-shirt Folding) outperforming OpenVLA-OFT by 22.2\%.
    \item[2)] Combined with \textbf{Tab.~\ref{tab:efficiency_real}}, SemanticVLA demonstrates substantial improvements in both training and inference. Notably, in ALOHA setup with 25 actions per chunk, SA-Coupler reduces action tokens per inference from 350 to 150, substantially cutting inference overhead.
    \item[3)] As visualized in \textbf{Fig.~\ref{vis1}}, SemanticVLA consistently completes complex instructions across different execution stages, demonstrating strong instruction-following capability and generalization in real-world scenarios.
\end{enumerate}

\begin{table}[t]
\caption{Ablation on sparsification ratio (Spf.Ratio) \& Comparison with FastV~\cite{fastv} and SliME~\cite{slime} under 8$\times$. ``\dag'' denotes reproduced results.}
    \label{tab:sparsification}
    \centering
    \footnotesize
    \setlength{\tabcolsep}{4pt}  
    \begin{tabular}{c|cccc}
    \toprule
    Spf.Ratio & \textbf{SR} $\uparrow$ & \textbf{FLOPs} $\downarrow$ & \textbf{Taining Cost} $\downarrow$ &\textbf{Latency} $\downarrow$ \\
    \midrule
    4 $\times$ & 97.7 & 3.28 T & 4.5 h & 0.093 s \\
    8 $\times$ & 97.7 & 2.37 T & 3.9 h & 0.089 s\\
    16 $\times$ & 95.8 & 1.93 T & 3.6 h & 0.087 s\\
    32 $\times$ & 92.0& 1.72 T& 3.5 h & 0.086 s\\
    \midrule
    \midrule
    FastV\dag  & 88.8  & 2.71 T  & - & 0.091 s\\
    SliME\dag & 85.6 & 2.71 T & 3.8 h & 0.089 s\\
    \bottomrule
    \end{tabular}
\end{table}

\subsection{4.3~~Ablation Studies}
\paragraph{Ablation on SD-Pruner.} As shown in \textbf{Tab.~\ref{tab:SD-Pruner}}, 1) SigLIP with ID-Pruner enables instruction-driven token pruning via language-supervised feature alignment, thus maximizing semantic density. 2) DINOv2 with SA-Pruner preserves global geometric structure through token aggregation, while injecting lightweight semantics through FiLM. Their targeted combination in SemanticVLA yields both semantic focusing and geometric perception, outperforming inverse or single configurations by 2.1\%–5.2\% in success rate.

\paragraph{Ablation on sparsification ratio.} \textbf{Tab.~\ref{tab:sparsification}} presents results under varying sparsification ratios $R \in \{4, 8, 16, 32\}$. The chosen setting $R = 8$ attains a 97.7\% success rate, offering a favorable trade-off between performance and efficiency. $R = 16$ yields a marginal 1.9\% drop and defines the SemanticVLA-Lite variant. In contrast, $R = 4$ retains redundancy, limiting speedup, while $R = 32$ discards critical semantic context, degrading performance. Furthermore, compared to other plug-and-play sparsification baselines at the same 8$\times$ compression level (e.g., FastV and SliME), SemanticVLA exhibits significantly better performance, underscoring that only instruction-aware pruning combined with structural preservation via HF-Fuser achieves Pareto-optimality in both performance and efficiency.

\paragraph{Ablation Study on HF-Fuser and SA-Coupler.} As shown in \textbf{Tab.~\ref{tab:ffp-datc}}, HF-Fuser and SA-Coupler provide complementary improvements across all tasks, with the largest gains in long-horizon tasks, highlighting the effectiveness of SemanticVLA’s vision-action co-design. 
Specifically, HF-Fuser improves success rates by hierarchically integrating fine-grained observation tokens from both visual encoders. SA-Coupler eliminates redundant action tokens and reduces overfitting in the action space, particularly under sparse visual inputs. Together, these modules operate at different token granularities and enhance cross-modal alignment, yielding synergistic improvements beyond additive contributions.

\begin{table}[t]
\caption{\textbf{Ablation Study on HF-Fuser (HF-F) and SA-Coupler (SA-C).}  ``\texttimes'' for SA-Coupler denotes the use of uncompressed 7-DoF action tokens.}
    \label{tab:ffp-datc}
    \centering
    \footnotesize
    \setlength{\tabcolsep}{5pt}  
    \begin{tabular}{cc|ccccc}
    \toprule
    \textbf{HF-F} & \textbf{SA-C} & \textbf{Spatial} & \textbf{Object}  & \textbf{Goal} & \textbf{Long} & \textbf{Overall} \\
    \midrule
     \texttimes & \texttimes & 95.2 &  96.0& 94.4 & 88.6 & 93.6 \\
     \checkmark & \texttimes & 96.8 & 97.4 & 95.6 & 92.4 & 95.6\\
     \texttimes &  \checkmark  & 95.6 & 96.4 & 95.2 & 89.2 & 94.1 \\
    \checkmark & \checkmark & \textbf{98.2} & \textbf{99.0} & \textbf{97.2} & \textbf{93.8} & \textbf{97.1}\\
    \bottomrule
    \end{tabular}
\end{table}

\subsection{4.4~~Qualitative Analysis}
\textbf{Fig.~\ref{vis1}} shows that the model consistently produces instruction-aligned action sequences with minimal deviation on three long-horizon real-world tasks. Each frame marks a key subgoal transition, demonstrating robustness in complex manipulation workflows. \textbf{Fig.~\ref{vis2}} demonstrates that ID-Pruner in SigLIP emphasizes both global action cues and local semantic anchors, while SA-Pruner in DINOv2 focuses on geometric structure, highlighting their complementary strengths in semantic and spatial grounding. These visualizations indicate that the synergy among SD-Pruner, HF-Fuser, and SA-Coupler produces interpretable intermediate representations and reliable downstream control, validating the effectiveness of Semantic-Aligned Sparsification and Enhancement in real-world execution.

\section{5~~Conclusion}
We present SemanticVLA, a novel framework for Semantic-Aligned Sparsification and Enhancement in robotic manipulation. By integrating Semantic-guided Dual Visual Pruner (SD-Pruner) for semantic-guided visual sparsification, Semantic-complementary Hierarchical Fuser (SH-Fuser) for cross-encoder semantic-structural fusion, and Semantic-conditioned Action Coupler (SA-Coupler) for modular action control, SemanticVLA achieves state-of-the-art task success with significantly reduced computational cost. Extensive evaluations on simulation and real-world tasks demonstrate its robustness, scalability, and efficiency. We hope this work offers new insights for advancing research in the embodied intelligence community.

\bibliography{supp}

\cleardoublepage
\appendix
We provide comprehensive supplementary material to support the methodology, implementation, and analysis introduced in the main paper on SemanticVLA. The content is organized as follows:
\begin{itemize}
    \item \textbf{Section A}\ref{sec:implementation} details the implementation of SemanticVLA, including component composition, parameter configurations, and training settings.
    \item \textbf{Section B}\ref{sec:simulation_benchmark} presents a complete breakdown of LIBERO benchmark task suites.
    \item \textbf{Section C}\ref{sec:real_world_setup} outlines the real-world experimental setup, covering task definitions, evaluation protocols, and data augmentation strategies to enhance generalization and robustness.
    \item \textbf{Section D}\ref{sec:quantitative_analysis} includes additional results from both simulation benchmarks and real-world evaluations.
    \item \textbf{Section E}\ref{sec:experimental_analysis} provides extended qualitative analyses, including execution visualizations in simulated and real environments, and attention maps of different visual tokens, and key language token selection.
    \item \textbf{Section F}\ref{sec:discussion} offers an in-depth discussion on the theoretical motivations behind SemanticVLA, current limitations, open research questions, and potential societal impacts and risks.
\end{itemize}



\section{A~~Implementation Details}
\label{sec:implementation}

\subsection{A.1~~Model Details}

For SemanticVLA, the number of tokens retained by the VL Mapping module in the SigLIP encoder and the number of aggregation tokens used in DINOv2 are both set to 32. This achieves an 8$\times$ compression rate of visual tokens. 

For SemanticVLA-Lite, both k and the number of aggregation tokens are set to 16, resulting in a 16$\times$ compression rate. Specifically, the input action placeholder is set to 1, which further reduces the number of action tokens processed by the LLM by a factor of 3. 

For both SemanticVLA and SemanticVLA-Lite, the LV Filtering module in SigLIP extracts $h = 5$ key semantic tokens. Additionally, DINOv2 and SigLIP have different numbers of layers, with Dinov2 having 24 and SigLIP having 27 layers. We select layers $\{4, 14, 24\}$ from DINOv2 and $\{3, 12, 21\}$ from SigLIP to represent shallow, intermediate, and deep layers for the two encoders, respectively. These selected layers are then used to establish cross-encoder interactions through the FFP module.

\subsection{A.2~~Training Details}

\paragraph{LIBERO Training Setup.} We use OpenVLA~\cite{openvla} as the backbone model and set the action chunk size to $K=8$. Fine-tuning is conducted using Low-Rank Adaptation (LoRA) with a rank of 64 and an $\alpha$ value of 128. The model is trained for 80K steps with a batch size of 128 and an initial learning rate of 5e-4. A linear warm-up strategy is applied during the first 2000 steps. After that, a cosine decay schedule is used to gradually reduce the learning rate to a final value of 1e-5. For ablation studies, we use a lower rank of 32 and a batch size of 64 to improve efficiency. Checkpoints are evaluated every 10K steps. The checkpoint with the best performance is selected for reporting.

\paragraph{Real-World Training Setup.} For the real-world experiments, we set the chunk size to $K=25$ and fine-tune OpenVLA using LoRA with a rank of 32 and an alpha value of 64. The model is trained with a batch size of 32 and adopts the same learning rate schedule as LIBERO training. Starting from step 60K, we evaluate checkpoints every 10K steps and report the best-performing checkpoint.


\begin{figure}[t]
\centering
\includegraphics[width=0.6\columnwidth]{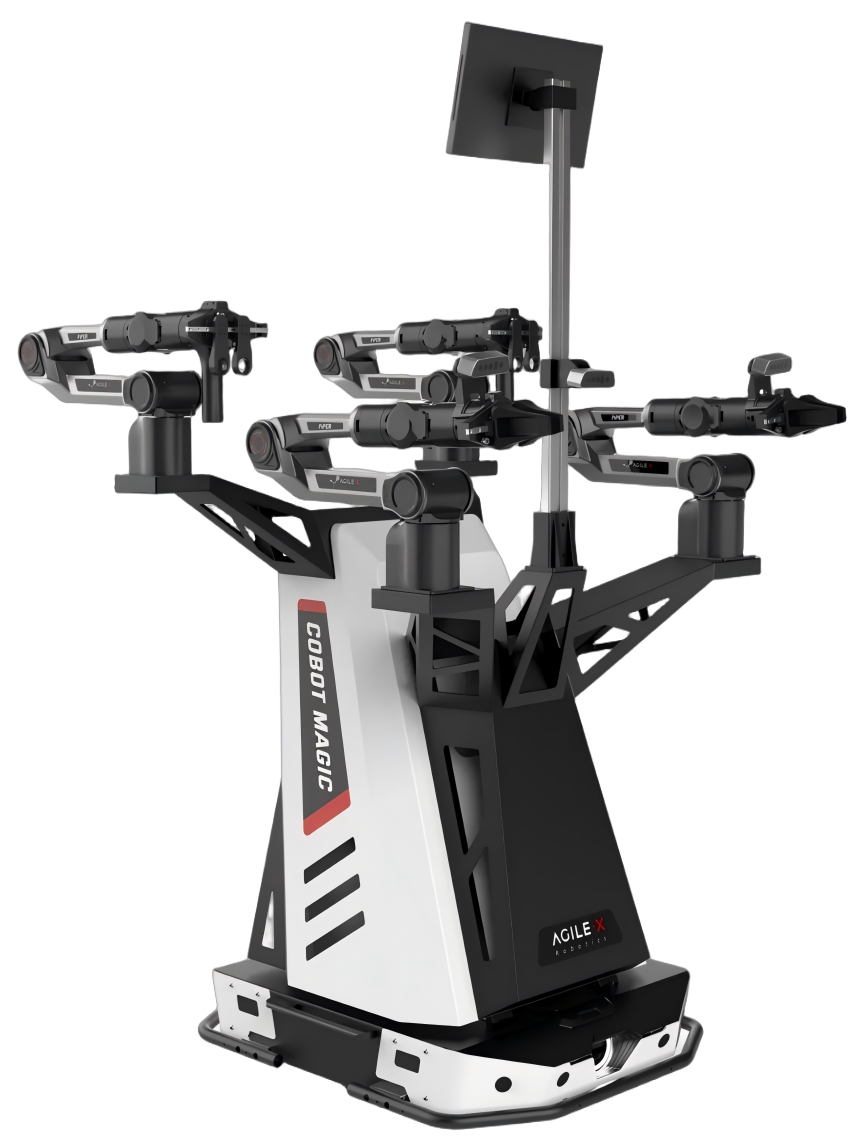}
\caption{AgileX Cobot Magic platform.}
\label{app1}
\end{figure}

\section{B~~Simulation Benchmark}
\label{sec:simulation_benchmark}

\subsection{B.1~~LIBERO Task Suites}
LIBERO (LIfelong learning BEnchmark for RObot manipulation) is a simulation-based evaluation platform designed to study lifelong learning and multi-task transfer in robotic manipulation. It consists of four structured task suites:
\begin{itemize}
    \item \textbf{LIBERO-Spatial} (10 tasks): Focuses on spatial reasoning, such as recognizing relative object positions. Object types are fixed, with variations only in spatial configurations, enabling assessment of spatial knowledge transfer.
    
    \item \textbf{LIBERO-Object} (10 tasks): Evaluates semantic understanding of diverse objects, such as selecting specific items for manipulation. Object positions remain constant, while categories vary, targeting generalization over declarative knowledge.
    
    \item \textbf{LIBERO-Goal} (10 tasks): Tests transfer of action sequences and goal-directed behaviors (procedural knowledge), such as placing or opening objects, with fixed layouts and object types.
    
    \item \textbf{LIBERO-Long} (10 tasks): Integrates complex and overlapping spatial, object, and goal specifications to assess long-term learning under highly mixed-task conditions.
\end{itemize}
LIBERO offers a well-structured and comprehensive benchmark for lifelong learning, supporting holistic evaluation of model performance, stability, and generalization.

\section{C~~Real-World Setup}
\label{sec:real_world_setup}

\subsection{C.1~~Robot Platform Setup}
As shown in \textbf{Fig.~\ref{app1}}, we validate real-world task performance on AgileX Cobot Magic platforms: Based on Stanford’s Mobile ALOHA project, this platform by AgileX Robotics integrates a differential-drive AGV base Tracer, dual-arm manipul~\cite{aloha}ators, and RGB-D sensors. It supports remote teleoperation and autonomous execution of complex tasks such as watering plants, cooking, and bimanual grasping.



\begin{table*}[t]
\caption{\textbf{Real-World Results (1).} Comparison of Task 1-3 and their subtask success rates on the Galaxea R1 Lite platform, with 15 trials conducted per task. The overall success rate (SR) reflects only the final subtask outcomes. ``\dag'' denotes our reproduced results.}
    \centering
    \footnotesize
    \setlength{\tabcolsep}{4pt}
    \begin{tabular}{l|cc|ccc|ccc|c}
    \toprule
    \textbf{Method} & \multicolumn{2}{c|}{\textbf{Object Placement}} & \multicolumn{3}{c|}{\textbf{Drawer Manipulation}} & \multicolumn{3}{c|}{\textbf{T-shirt Folding}} & \textbf{Overall} \\
    & Bear $\rightarrow$Plate & +Raccoon$\rightarrow$Bowl & Open  & +Place & +Close & Step 1 & +Step 2 & +Step 3 & SR\\
    \midrule
    \multicolumn{10}{c}{\textbf{Galaxea R1 Lite platform}} \\
    \midrule
    PD-VLA\dag~\cite{pdvla} & 11/15 & 10/15 & 10/15 & 8/15 & 7/15 & 9/15 & 8/15 & 6/15 & 51.1\%\\
    OpenVLA-OFT\dag~\cite{oft} & 12/15 & 10/15 & 11/15 & 9/15 & 7/15 & 10/15 & 8/15 & 7/15& 53.3\%\\
    \textbf{SemanticVLA }& \textbf{14/15} & \textbf{13/15} & \textbf{13/15} & \textbf{11/15} & \textbf{10/15} & \textbf{13/15} & \textbf{12/15} & \textbf{10/15}& \textbf{73.3\%} \\
    \bottomrule
    \end{tabular}
\end{table*}

\begin{table*}[t]
\caption{\textbf{Real-World Results.} Comparison of Task 4-5 and their subtask success rates on the AgileX Cobot Magic platform, with 15 trials conducted per task. The overall success rate (SR) reflects only the final subtask outcomes. ``\dag'' denotes our reproduced results. The results demonstrate the robustness and strong performance of SemanticVLA across different robotic platforms.}
    \label{task1-3}
    \centering
    \footnotesize
    \setlength{\tabcolsep}{10pt}
    \begin{tabular}{l|cc|c|c}
    \toprule
    \textbf{Method} & \multicolumn{2}{c|}{\textbf{Task 4}} & \textbf{Task 5} & \textbf{Overall} \\
    & Orange Cube$\rightarrow$Plate & +Big Cube$\rightarrow$Bowl & Left Cube$\rightarrow$Plate  & SR\\
    \midrule
    \multicolumn{5}{c}{\textbf{AgileX Cobot Magic platform}} \\
    \midrule
    PD-VLA\dag~\cite{pdvla}  & 10/15 & 7/15 & 10/15& 56.7\%\\
    OpenVLA-OFT\dag~\cite{oft}  & 10/15 & 8/15 & 10/15& 60.0\%\\
    \textbf{SemanticVLA}  & \textbf{12/15} & \textbf{10/15} & \textbf{13/15}& \textbf{76.7\%} \\
    \midrule
    \multicolumn{5}{c}{\textbf{Galaxea R1 Lite platform}} \\
    \midrule
    PD-VLA\dag~\cite{pdvla}  & 10/15 & 8/15 & 10/15& 60.0\%\\
    OpenVLA-OFT\dag~\cite{oft}  & 10/15 & 8/15 & 10/15& 60.0\%\\
    \textbf{SemanticVLA}  & \textbf{12/15} & \textbf{11/15} & \textbf{13/15}& \textbf{80.0\%} \\
    \bottomrule
    \end{tabular}
\end{table*}

\begin{table*}[t]
\caption{\textbf{Efficiency Results in Real-World.} SemanticVLA-Lite and SemanticVLA achieve the highest efficiency and the best trade-off between efficiency and performance, respectively. ``\dag'' denotes our reproduced results. Z and H indicate the number of visual input tokens and initialized action tokens. Throughput refers to the number of actions predicted per second during inference.}
    \label{tab:efficiency_real}
    \centering
    \footnotesize
    \setlength{\tabcolsep}{5.2pt}
    \begin{tabular}{l|cccccc}
    \toprule
    \textbf{Method} & \textbf{Z \& H tokens} $\downarrow$ & \textbf{FLOPs} $\downarrow$ & \textbf{Training Cost} $\downarrow$ & \textbf{Latency} $\downarrow$ & \textbf{Throughput} $\uparrow$ & \textbf{LIBERO SR} $\uparrow$ \\
    \midrule
    OpenVLA\dag~\cite{openvla} & 256 \& 7 & 16.3 T  & 12.8 h  & 0.552 s   & 1.8 Hz & -  \\
    OpenVLA-OFT\dag~\cite{oft} & 256 \& 7 & 16.2 T  & 13.7 h  & 0.321 s  & 77.9 Hz  & 55.6\%  \\
    PD-VLA\dag~\cite{pdvla} & 256 \& 7  &16.3 T  & 12.8 h     & 0.350 s  & 71.4 Hz   & 51.1\%  \\
    \midrule
    \textbf{SemanticVLA-Lite} & 16 \& 3 & 3.1 T   & 2.8 h  & 0.168 s  & 148.8 Hz  & 62.2\% \\
    \textbf{SemanticVLA}  & 32 \& 3 & 5.0 T  & 3.3 h    & 0.183 s  & 136.6 Hz  & 77.8\% \\
    \bottomrule
    \end{tabular}
\end{table*}

\subsection{C.2~~Task Description \& Evaluation Protocol}
In addition to Tasks 1–3 reported in the main paper, we also trained and evaluated Tasks 4–5. A total of 60, 60, 45, 60, and 45 expert demonstrations were collected via human teleoperation for Tasks 1 through 5, respectively. Each task was evaluated over 15 trials. SemanticVLA demonstrates consistently strong performance across all tasks. Detailed results are presented in \textbf{Appendix~D}.
\begin{itemize}
    \item \textbf{Task 1}: \textit{``Put the bear into the plate, and then put the raccoon into the bowl.''} \\
    This task involves sequential object placement conditioned on category-specific visual recognition. The robot must correctly identify and manipulate two distinct animal-shaped objects—a bear and a raccoon—and place them into two designated receptacles. Success depends on accurate object grounding and adherence to temporal order in execution.
    
    \item \textbf{Task 2}: \textit{``Open the drawer, place the toy into the drawer, and then close it.''} \\ 
    This task evaluates articulated object manipulation and temporal coordination across three discrete actions. The agent is required to interact with a drawer mechanism, inserting an object before restoring the drawer to its original closed state. The task integrates rigid-body interaction, containment reasoning, and state restoration.
    
    \item \textbf{Task 3}: \textit{``Fold the T-shirt.''} \\ 
    A deformable object manipulation task that necessitates precise control over flexible materials. The robot is expected to perform a series of folding operations on a T-shirt, assessing its capacity for compliant motion planning, shape estimation, and fine-grained dual-arm coordination under deformation dynamics.
    
    \item \textbf{Task 4}: \textit{``Pick the orange cube into the plate, and then pick the big cube into the bowl.''} \\  
    This task tests the agent’s ability to resolve multi-attribute references involving both color and size. It requires grounding of compositional language instructions and execution of two pick-and-place operations differentiated by object attributes and target locations.
    
    \item \textbf{Task 5}: \textit{``Pick the left cube into the plate.''} \\ 
    A spatially grounded object selection task emphasizing egocentric perspective-taking. The robot must disambiguate multiple visually identical objects based on relative positioning and execute a single pick-and-place action toward the specified container.
\end{itemize}


\section{D~~Extensive Quantitative Analysis}
\label{sec:quantitative_analysis}

\subsection{D.1~~Real-World: Additional Tasks}

As shown in \textbf{Tab.~\ref{task4-5}}, we further introduce two additional real-world tasks. While Tasks 1–3 already cover key manipulation types such as object classification, articulated-object interaction, and deformable object handling, Tasks 4-5 are designed to assess the model’s capabilities in multi-attribute parsing and spatial reasoning. In Task 4, which involves two subtasks (``place the orange cube into the plate'' and ``place the large cube into the bowl''), SemanticVLA achieves higher success counts than baselines on both platforms, demonstrating a strong understanding of compositional instructions involving attributes like color and size. Task 5 further evaluates spatial reasoning and relative position understanding, where SemanticVLA achieves a high success rate of 13/15 on both platforms, substantially outperforming other methods.

\subsection{D.2~~Real-World: Efficiency Results}
As illustrated in Tab.~\ref{tab:efficiency_real}, SemanticVLA demonstrates even more pronounced efficiency advantages on the real-world ALOHA bimanual manipulation tasks. It outperforms OpenVLA and other efficient baseline models significantly in terms of training efficiency (FLOPs, cost) and inference efficiency (latency, throughput).

\begin{figure*}[!t]
\centering
\includegraphics[width=1.0\textwidth]{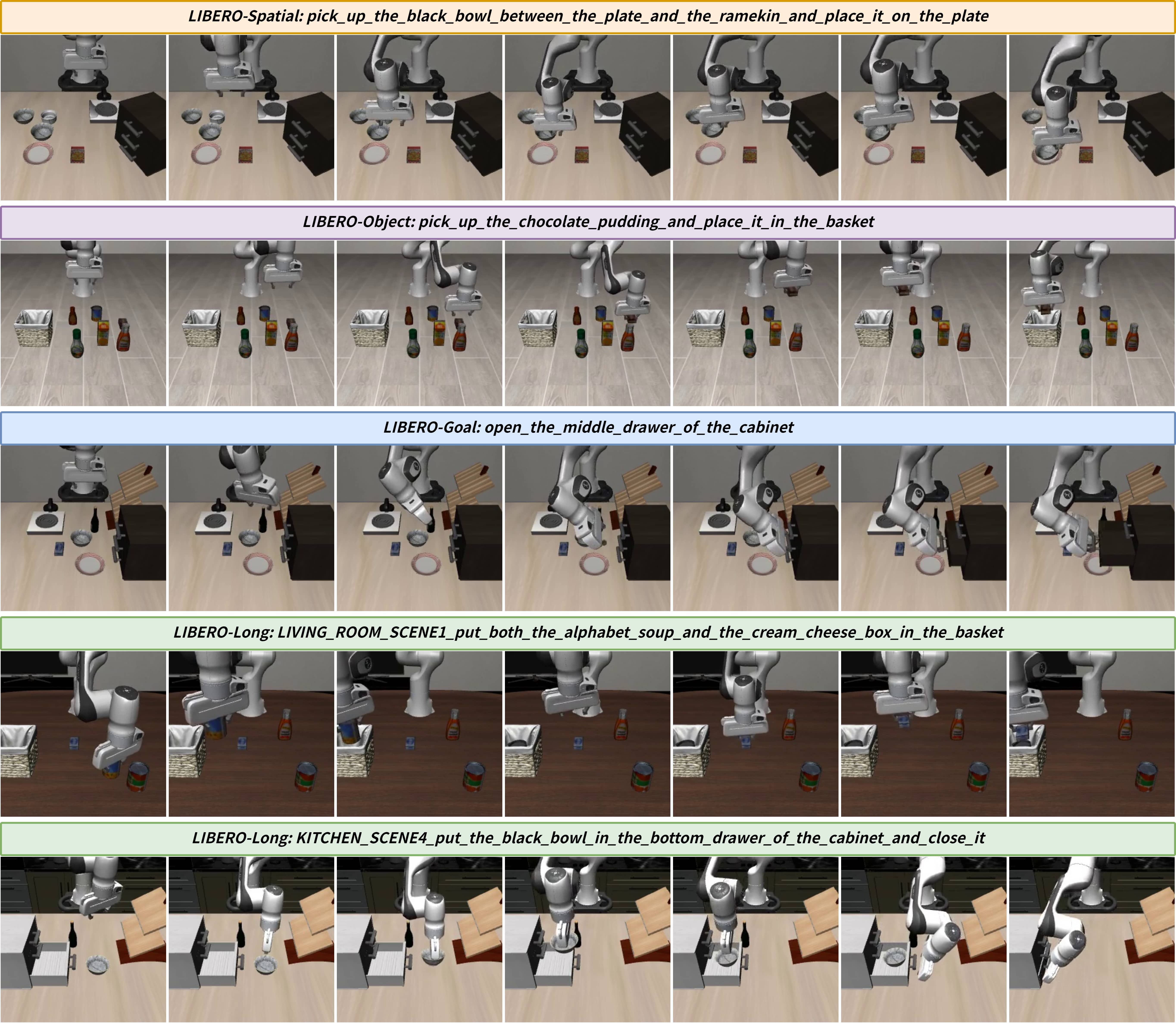}
\caption{Simulation execution visualizations of SemanticVLA across the four LIBERO task suites. The model demonstrates strong capabilities in grounding geometric relations, distinguishing fine-grained visual differences, executing diverse manipulation tasks, and handling complex multi-step procedures.}
\label{vis-libero}
\end{figure*}

\begin{figure*}[!t]
\centering
\includegraphics[width=1.0\textwidth]{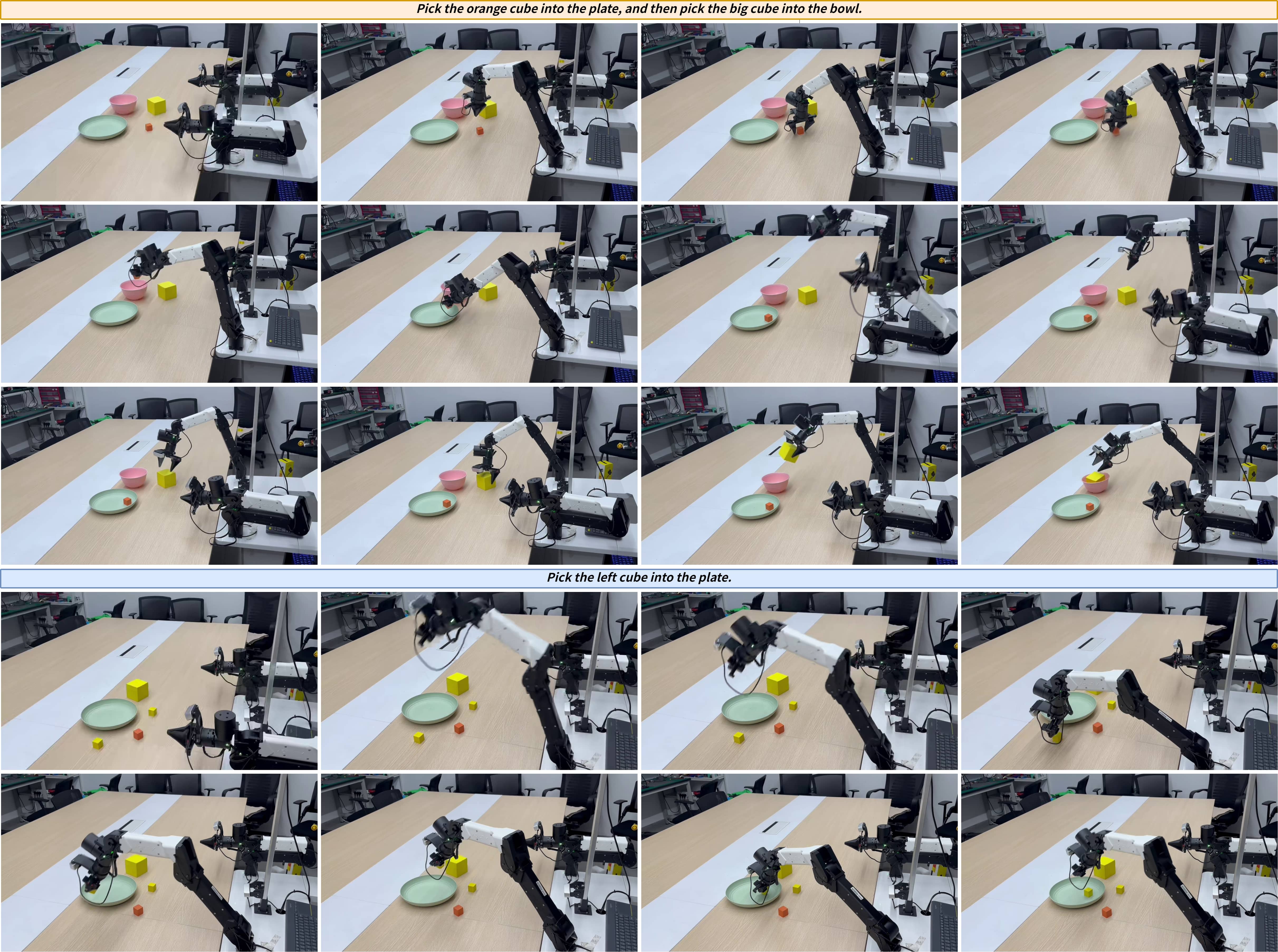}
\caption{Third-person real-world execution of Task 4 and Task 5. In Task 4 (top), SemanticVLA correctly grounds compositional attributes (“orange”, “big”) to complete two sequential placements. In Task 5 (bottom), it accurately selects the leftmost cube despite visual ambiguity, showing strong spatial reasoning and target disambiguation. Each frame highlights a critical stage of perception and action during task execution.}
\label{task4-5}
\end{figure*}

\begin{figure*}[!t]
\centering
\includegraphics[width=1.0\textwidth]{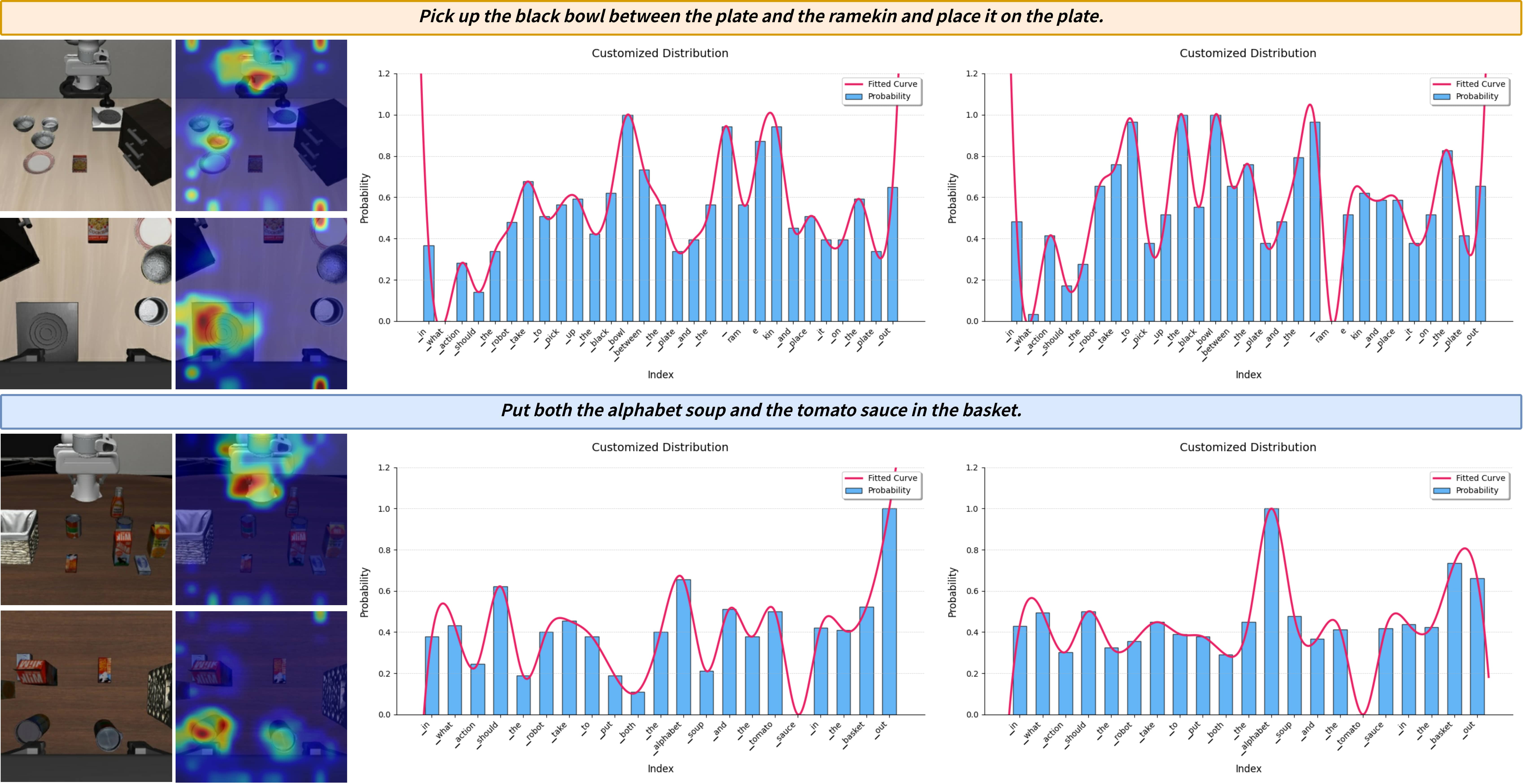}
\caption{Visualization of semantic cue words in the Vision-to-Language (VL) mapping process. The left panel shows RGB observations and attention heatmaps from top-down and egocentric views. The right panel presents normalized attention distributions over language tokens, along with fitted curves (blue bars for token weights, red line for trend fitting). Attention peaks highlight key cue words leveraged by the model during execution, illustrating the alignment between visual regions and linguistic semantics.}
\label{word}
\end{figure*}

\section{E~~Extensive Qualitative Analysis}
\label{sec:experimental_analysis}

\subsection{E.1~~Simulation: Execution Trajectories}
We visualize SemanticVLA’s execution across the four task suites in the LIBERO benchmark, as shown in \textbf{Fig.~\ref{vis-libero}}. In the Spatial suite, SemanticVLA accurately localizes key geometric anchors such as ``the black bowl between the plate and the ramekin,'' validating the effectiveness of semantic–geometric complementary sparsification. In the Object suite, it distinguishes fine-grained differences among visually similar objects (e.g., chocolate pudding) under challenging, cluttered settings. In the Goal suite, SemanticVLA executes distinct manipulation tasks within scenes structurally similar to the Spatial suite, with minimal redundant adjustment throughout the pick-and-place process. In the most challenging Long suite, it successfully completes complex long-horizon tasks across diverse scenarios, demonstrating robust compositional and temporal reasoning.

\subsection{E.2~~Real-World: Execution Trajectories}
We present third-person visualizations of Task 4 and Task 5 in \textbf{Fig.~\ref{task4-5}}, illustrating key visual observations and action responses during the execution of SemanticVLA. In Task 4, the model first parses compositional attributes in the instruction, including color (orange) and size (big). It successfully identifies and places the orange cube onto the plate, then transitions to locate and manipulate the large cube. This reflects precise understanding of combined attributes and efficient action planning. In Task 5, despite visual similarity among objects, SemanticVLA selects the leftmost cube based on its egocentric view, demonstrating strong spatial referencing and relative position reasoning capabilities.

\subsection{E.3~~Cue Word Selection}
To gain deeper insight into SemanticVLA’s cue word selection process, we visualize the Vision-to-Language (VL) mapping in \textbf{Fig.~\ref{word}}, focusing on two representative tasks involving spatial referencing and multi-object grounding. The left side displays RGB observations from egocentric and top-down views, along with the attention heatmap corresponding to a single V-to-L token. The right side presents normalized attention probabilities across language tokens with fitted curves, revealing key semantic anchors. Notable observations include:
\begin{itemize}
    \item We identify five cue word tokens that directly guide the compression of 256 original patch tokens into five V-to-L tokens. As shown by the heatmaps (one example illustrated), these compressed tokens effectively summarize global action-relevant cues within the scene.
    
    \item The selected cue words exhibit sharp attention peaks in the fitted curves, indicating their critical role in instruction interpretation. These tokens are extracted through the full end-to-end action prediction process, and their cosine similarity with patch tokens yields globally representative visual embeddings.
    
    \item Interestingly, as shown in Figure~5 of the main paper, the global attention maps derived from different cue word tokens are broadly consistent while preserving subtle semantic variations. This observation suggests that selecting a small number of tokens (e.g., 3-10) is sufficient to represent the full scene context effectively.
\end{itemize}
SemanticVLA achieves robust cross-modal alignment by automatically identifying and attending to key semantic cue words. This language-driven attention mechanism plays a central role in enabling precise instruction grounding, perceptual disambiguation, and action planning in complex manipulation tasks.

\section{F~~Discussion}
\label{sec:discussion}

\subsection{F.1~~Supplementary Details on the Motivation}
Human behavior shows that seeing is closely coupled with doing. Vision is not only a perceptual input but a core enabler of reasoning and creativity. It allows agents to perceive objects, infer structure, and plan actions. With the rise of spatial intelligence, the idea that ``creativity is fundamentally visual'' reflects the full cognitive loop from perception to execution. SemanticVLA exemplifies this principle. Its creativity lies not in generating content, but in efficiently grounding language in perception and executing precise actions in complex environments. This is achieved through three vision-guided modules:
\begin{itemize}
    \item ID-Pruner selects semantically relevant regions from cluttered scenes based on language, enabling goal-aware visual sparsification and semantic reasoning.
    
    \item Dense-Fuser aligns visual features with semantic anchors, preserves spatial structure, and supports scene understanding and goal-directed planning.
    
    \item SA-Coupler compresses continuous 7-DoF trajectories into type-coupled representations, supporting efficient execution of long-horizon tasks.
\end{itemize}
Creativity here is expressed through perception-driven decision-making, enabling agents to act meaningfully based on what is seen.


\subsection{F.2~~Limitations \& Future Work of SemanticVLA}
Despite SemanticVLA’s strong performance in both simulation and real-world robotic manipulation tasks, several limitations persist. While the framework achieves computational efficiency through visual token and action compression, it does not yet incorporate active perception or memory mechanisms. These components are important for effective task execution in long-horizon and partially observable scenarios. Furthermore, extending the system’s language understanding capabilities to handle more compositional, abstract, or dialogue-driven instructions remains a significant challenge.

Future work will focus on: 1) incorporating reinforcement learning or meta-learning to enable more adaptive action prediction strategies; 2) augmenting SemanticVLA with visual memory and temporal reasoning modules to support persistent long-horizon execution; and 3) integrating interactive language grounding, such as dialogue-based or corrective feedback mechanisms, to enhance the system’s usability in open-world environments.

\subsection{F.3~~Broader Impact \& Potential Risks}
SemanticVLA aims to advance efficient and generalizable vision-language-action (VLA) systems for robotic manipulation. Through instruction-aware sparsification and modular action reasoning, it enhances robots’ ability to interpret human instructions and execute tasks with greater precision and efficiency. This enables potential benefits in assistive technologies, automation of repetitive labor, and improved accessibility in everyday environments.

However, deploying such models in broader real-world settings entails potential risks. Misinterpretation of ambiguous or biased instructions may result in unintended or unsafe behaviors, particularly in open-world or safety-critical contexts. Moreover, reliance on pre-trained vision-language models can perpetuate societal biases embedded in the training data. Responsible deployment of SemanticVLA requires systematic evaluation across diverse scenarios to ensure fairness, reliability, and practical applicability.

\end{document}